\pdfoutput=1

\documentclass[11pt]{article}

\usepackage{acl}

\usepackage{times}
\usepackage{latexsym}

\usepackage[T1]{fontenc}

\usepackage[utf8]{inputenc}

\usepackage{microtype}
\usepackage{graphicx}
\usepackage{rotating, multirow, makecell}
\usepackage{enumitem}
\usepackage{stfloats}

\usepackage{xcolor}

\newcommand*{\rot}[1]{\multicolumn{1}{c}{\turnbox{60}{\multirowcell{2}[0pt][l]{#1}}}}

\title{Infusing Commonsense World Models with Graph Knowledge}

\author{Alexander Gurung\thanks{\hspace{.5em}Work done during a Meta AI residency} \\
  Georgia Institute of Technology
  \\\And
  Mojtaba Komeili\\
  Meta AI\\
  \\\AND
  Arthur Szlam\thanks{\hspace{.5em}Work done while at Meta AI}\\
  Meta AI\\
  \\\And
  Jason Weston \\
  Meta AI\\
  \\\And
  Jack Urbanek\\
  Meta AI}

\begin{document}
\maketitle
\begin{abstract}
While language models have become more capable of producing compelling language, 
we find there are still gaps in maintaining consistency, especially when describing events in a dynamically changing world. We study the setting of generating narratives in an open world text adventure game, where a graph representation of the underlying game state can be used to train models that consume and output both grounded graph representations and natural language descriptions and actions. 
We build a large set of tasks by combining crowdsourced and simulated gameplays with a novel dataset of complex actions in order to to construct such models.
We find it is possible to improve the consistency of action narration models by training on graph contexts and targets, even if graphs are not present at test time. This is shown both in automatic metrics and human evaluations. We plan to release our code, the new set of tasks and best performing models.

\end{abstract}

\section{Introduction}

While recent work has found the genre of text-adventure games to be a fruitful source for collecting rich context-dependent datasets, these games often suffer from one of two problems. Some rely on a rule-based game engine and thus have high fidelity but extremely restricted or formulaic responses, such as in TextWorld \citep{coteTextWorldLearningEnvironment2018}, and JerichoWorld \citep{ammanabroluModelingWorldsText2021}. Others use a generative model and suffer from failures in consistency and commonsense reasoning in order to handle an open set of actions, such as can be observed in AI Dungeon 2 \citep{aidungeon}. We aim to combine the benefits of these two approaches while mitigating the drawbacks by grounding the generations with a knowledge graph and training our model on novel grounding tasks. 

We focus on the rule-based text-adventure game environment LIGHT \citep{urbanekLearningSpeakAct2019}. LIGHT is a collection of text adventure game settings, elements, and in-game dialogues, as well as a functioning text adventure game engine based on a graph state. It includes 100s of locations, over 1000 characters and over 500k recorded in-game dialogue utterances and interactions in a rich open world, that can all be used for training models.

We use the LIGHT game engine to provide the game state and known transitions, and additionally collect entirely novel interactions not covered by the original engine. Leveraging these we aim to show how fine-tuning on grounding tasks can produce language models capable of understanding the relationships between actions and their environment in the real-world.

We first represent the internal game state using a graph, thus providing explicit information on details such as character and item attributes that we can include in a model's context during training. Following \citet{liuKGBARTKnowledgeGraphAugmented2021}, this representation takes the form of triples describing a relationship as \textit{<Item, Edge, Value>}. 
We then construct a dataset based on game locations which were
collected via crowdworkers in order to 
improve a model's internal representation of an environment. 
By removing elements of the graph or context and asking
the model to predict the missing elements, we hope a model learns what \textit{could} or \textit{should} be in a room but simply is not stated. We follow a similar procedure for creating tasks concerning game states and narrations following actions using a dataset of game playthroughs.

\begin{figure*}[!th]
    \centering
    \includegraphics[width=130mm]{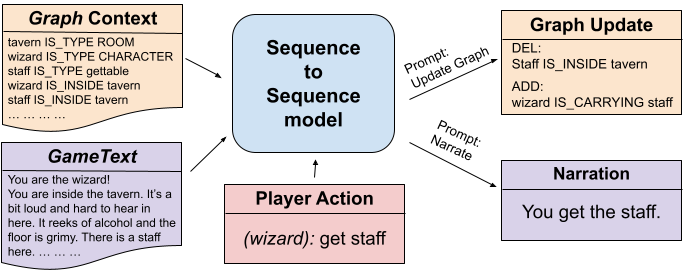}
    \caption{
    \textbf{World models with graph knowledge}: we train a sequence to sequence model with both text narration and graph representation formats to infuse commonsense into the model. Even if graph information is not available in the context at inference time, such models can still outperform models trained without graph knowledge.  
    }
    \label{fig:main_experiment}
\end{figure*}

We further collect a new dataset, \textsc{UseEvents}, containing multi-object actions and their corresponding narrations and effect on the world. These complex interactions are not commonly found in rule-based text-adventure game engines. By including this new dataset in our tasks we hope to improve the adaptability of our models and reduce its reliance a limited set of coded actions.

We then fine-tune a sequence to sequence model \citep{seq2seq} on various combinations of the above datasets. Our setup  is summarized in Figure \ref{fig:main_experiment}. After selecting the best performing models via automatic metrics, we find that the best model according to human evaluators was trained with graph grounding, even when it is not present during evaluation. We further analyze the qualitative differences between these models. We plan to make our entire setup, code and best models publicly available.


In short, our contributions are as follows: 
\begin{enumerate}
\itemsep0em 
    \item We introduce a graph-text representation of LIGHT's internal game state, and use this representation to build a variety of action and environment tasks with the goal of grounding a model's understanding of its world.
    \item We collect a new dataset of \textsc{UseEvents} to expand to an open-world set of actions and narrations.
    \item We show that a combination of the above can outperform non-grounded language model generation methods as measured by both automatic metrics and human evaluations.
\end{enumerate}

\section{Related Work}

\subsection{Text-Adventure Games}

Text-adventure games like TextWorld \citep{coteTextWorldLearningEnvironment2018}, JerichoWorld \citep{ammanabroluModelingWorldsText2021}, and LIGHT \citep{urbanekLearningSpeakAct2019} have all found success as testbeds for progress in reinforcement learning (RL), language understanding and generation. TextWorld and JerichoWorld focus primarily on RL interactions in classic text adventure games, and LIGHT more directly focuses on language by incorporating dialogue in a multi-agent setting. Collecting quality data and evaluating RL models can be easier in these rich and simulatable environments, and prior work has demonstrated potential avenues for improvements in dialogue agents \citep{Prabhumoye2020ILY}, and narrative generations \citep{ammanabrolu-etal-2019-toward} in this domain.

\subsection{Commonsense Reasoning}

Commonsense reasoning research like that of CommonGen \citep{Lin2019CommonGenAC}, SWAG/HellaSWAG \cite{Zellers2018SWAGAL, Zellers2019HellaSwagCA} has traditionally relied on crowdsourced datasets collected from online data or from crowdsourced scenarios. ATOMIC \citep{sapATOMICAtlasMachine2019} extended this methodology by proposing \textit{if-then} relationships between events, personas, and mental states. However, in the real-world contextual information about one's environment is often crucial in predicting the result of an action.

To better represent this grounded form of commonsense reasoning, PIGLeT \citep{zellersPIGLeTLanguageGrounding2021} operated in a simulated 3D environment THOR by grounding language models (which communicate with physical dynamics models) via action triples. We use a related methodology in a text-adventure game setting where,  with the addition of  environment-specific grounding tasks, we can improve over traditional commonsense reasoning approaches.

\section{Grounding with  Graph Knowledge}

\subsection{Graph Representation}

Internally in the LIGHT text adventure game engine, similarly to other game engines, a given location is represented by nested objects and characters each with their own attributes. This gives the LIGHT context graph a tree structure. This structure is then used to generate a text representation of the state. However, this transformation is lossy and does not contain all of the details that encode the full game state. We will call this context provided to a player during gameplay the $GameText$.

In order to provide language models with an easily interpretable and complete representation of the game engine graph structure, we flatten the graph and represent all relationships as triples. 
For example, a coin inside of a box would be represented by \textit{<COIN, IS\_INSIDE, BOX>} and a wizard's description would be represented by \textit{<WIZARD, HAS\_DESCRIPTION, The wizard is wearing a pointy blue hat.>}. We refer to a newline-separated list of these triples as the $Graph$. 

When training models, we can build tasks that provide both the $GameText$ and the $Graph$ as context, alongside a task-specific prompt. Tasks that require modifying the graph by adding or removing tuples have an additional \textit{ADD: } or \textit{DEL: } token prepended to each update line to signify the type of graph modification. For example, after executing "wizard get staff" the label would have both "ADD: wizard IS\_CARRYING staff" and "DEL: staff IS\_INSIDE room". A complete example can be seen in Appendix Figure \ref{fig:graph_breakdown}. 

We find this graph representation to be easy to construct, interpret, and most importantly, for the models to learn from. However, one limitation of this approach is its dependence on the size of the context as particularly complex locations or long game histories can easily become larger than the context allows of a standard language model. In almost all cases concatenating the most recent information was sufficient for our purposes, but future work could for example use a more complex retrieval mechanism to resolve this issue.

\subsection{Dataset Construction}

\subsubsection{Action Tasks}

The first group of tasks we build are Action Tasks, whose goal it is to improve the model's understanding of how actions change the environment. These tasks are often the most difficult and most important to get right from a user's perspective, as the possible set of actions and consequences is very large and incorrect predictions can dramatically reduce the playability of the game. We define two classes of Action task, where we provide both the $GameText$ and the $Graph$ as context in each.

\textbf{Graph Update Tasks} ask the model to predict changes to the game's graph (as represented by our graph-text triples) given a character attempting to perform an action. We produce this task by partially rolling out episodes from the initial LIGHT dialogues database. We then set the next action from the dialogue as the target (for \textbf{GameActions} tasks) or from a random set (for \textbf{SelfPlay} tasks). We then use the LIGHT engine to compute the next graph state. Then we
use the game state both before and after the action to compute the difference caused by the action. We call this change a Graph Update, and it serves as the label for the task. By learning this task we intend to ground a model's understanding of causality into specific changes, e.g. an item changing owners or an item's attributes changing.

\textbf{Narration Tasks} ask a model to provide a rich text description of the result from a character's action. This is the description a user would see when playing the game, and is also the subject of our human evaluations. A traditional action-to-description model would be trained exclusively on this task. However, in this work we explore whether simultaneously learning grounding objectives  can improve performance on this task as well.

\subsubsection{Environment Tasks}

The second group of tasks are Environment Tasks, whose goal is to ground a model's understanding of the environment itself. We create these tasks by randomly removing elements from known graphs and asking the model to predict the missing elements using prior game rounds and the rest of the graph as context.

\textbf{Element Tasks} ask the model to provide a new element of that class using our graph-text triples. For example, we might ask a model to predict a character it believes could reasonably exist in this location but is not in the given context, or to predict an item that may be inside a container, again given other relevant context. As, for example, the placement of a hat on a character or inside other objects is important for understanding whether an action using it is possible, this related task is intended to teach the model to focus on the presence and location of elements.

\textbf{Attribute Tasks} ask the model to provide attributes for elements in the room at the current time step, like \textit{<box, IS\_CONTAINER, true>} for attributes in the LIGHT data model and \textit{<torch, HAS\_ATTRIBUTE, burnable>} for arbitrary attributes. We hypothesize that these tasks will help the model pay attention to underlying attributes of objects that may be relevant to future actions. For example, you may eat an orange or store a gold coin in a jar, but you are unlikely to eat a jar or store a gold coin in an orange. Some attributes may change after actions take place, for instance a torch would lose the attribute \textit{Lit} after the action 
\textit{Extinguish torch in lake}, so the model must also learn to look at the prior game history to fill in missing information.

\subsubsection{\textsc{UseEvents} Dataset}

\begin{table*}[!ht]
    \centering
    \small
    \begin{tabular}{|l |r | r |p{75mm}|}
    \hline
    \textbf{Action} & \textbf{Phrase} & \multicolumn{2}{p{95mm}|}{tie rope to wood stake}\\
    \cline{2-4}
    & \textbf{Narration} & \multicolumn{2}{p{95mm}|}{You tie the rope to each end of the wood stake. You can carry it on your back now.}\\
    \cline{2-4}
    & \textbf{Alternate} & \multicolumn{2}{p{95mm}|}{You tie the rope to each end of the stake and sling it across your back.}\\
    \cline{2-4}
    & \textbf{External} & \multicolumn{2}{p{95mm}|}{\{actor\} ties a rope to each end of a sharpened wooden stake and slings it across their back.}\\
    \hline
    \hline
    \textbf{Initial} & \textbf{Primary} & name & rope \\
    \cline{3-4}
    \textbf{Objects} & & description & a small length of tough rope \\
    \cline{2-4}
    & \textbf{Secondary} & name & sharpened wooden stake \\
    \cline{3-4}
    & & description & The wood stake has a sharp end, it looks really dangerous. It looks new and its made out of a snakewood tree. \\
    \hline
    \hline
    \textbf{Final} & \textbf{Name} & \multicolumn{2}{p{95mm}|}{sharpened wooden stake} \\
    \cline{3-4}
    \textbf{Objects} & Description & \multicolumn{2}{p{95mm}|}{The wood stake has a sharp end. It's new and made out of a snakewood tree. It has a piece of rope tied to each end, forming a sling.} \\
    \cline{3-4}
    & Location & \multicolumn{2}{p{95mm}|}{Worn by \{Actor\}} \\
    \cline{3-4}
    & Attribute Changes & \multicolumn{2}{p{95mm}|}{+Wearable} \\
    \hline
    
    \end{tabular}
    \caption{Example from the newly collected \textsc{UseEvents} dataset. Annotators were separately asked to create an action and then to ground that action. We can then use the complete set of groundings to know both the prerequisite and resultant graph states for a given action, as well as expected usage phrases and narrations.}
    \label{tab:use_event1}
\end{table*}

An inherent limitation of game engine-based datasets is their reliance on the action set of the game engine. As a result, the space of possible actions is restricted to the actions manually implemented in the codebase. One of the clearest examples of this restriction is in multi-object actions like \textit{roast fish over firepit}, which does not exist in the game engine action set. Another result of this restriction is the formulaic nature of narrations created via playthroughs. 

To ensure that our models are able to handle more complex actions and produce interesting narrations we collect a new dataset of \textsc{UseEvents}. Each UseEvent is an action with the base form \textit{use x with y} complete with: 

\begin{itemize}
\itemsep0em 
    \item a natural rephrasing of the base form
    \item constraints for this action
    \item attributes that change after this action
    \item a human-written narration of the action taking place
\end{itemize}

We crowdsource the dataset in stages, reaching a total of 10,000 fully defined \textsc{UseEvents}. Our staged approach allows us to have diverse and interesting events with complex grounding. One example can be seen in Table \ref{tab:use_event1}, with additional examples in Appendix \ref{sec:more-use-events}. A complete discussion of the collection process can be found in Appendix \ref{sec:use-event-collection}.

Using the information collected we can simulate these actions in the LIGHT game engine,  inserting the objects and constraints as necessary. We incorporate these simulations during training for the Action Tasks to improve narration quality and the models' understanding of action consequences, but do not incorporate these simulations in Environment Tasks as the inserted objects/constraints may not make logical sense in context.

\subsubsection{Dataset and Task Statistics}

We include the complete set of dataset statistics for the training splits of our datasets in Table \ref{tab:dataset_stats}. This displays details for the input and labels for each of our tasks. The \textbf{GameActionsNarration}, \textbf{SelfPlayActionsNarration}, \textbf{InvalidSelfPlayNarration}, and \textbf{UseEventActionsNarration} tasks comprise the \textbf{Narration Tasks} . The \textbf{GameActions}, \textbf{SelfPlayActions}, \textbf{InvalidSelfPlay}, and \textbf{UseEventActions} tasks comprise the \textbf{Graph Update Tasks}. All remaining tasks are part of the \textbf{Environment Task}  set.

Here we can see that \textsc{UseEvents} narrations for instance are much longer (24.83 tokens) than Game Action narrations (6.81 tokens). 

We additionally note that the \textsc{UseEvents} tasks have 500 additional examples in each of a \textit{valid}, \textit{test}, and \textit{unseen test} split. In this case, we define the ``unseen test'' split to be one where neither of the objects that appear in the interaction are present anywhere in the \textit{train} split.

\begin{table*}[!ht]
\centering
\small
\begin{tabular}{l l  r | r | r | r | r }
\textbf{Task} & &  Av. Length & Tokens & Unique Tokens & Unique Utterances & Utterances\\
\hline
{\textbf{AddCharacterCarrying}} &
input &  1722 & 7067606 & 15648 &  4105 & 4105\\
& labels &   11.96  &    49102 & 1550 &   2669 & 4105\\
\hline
{\textbf{AddCharacterDescription}} &
input  &  1423  &  20148798 &   11658 &  14123 & 14156\\
& labels &  16.53  & 234004 &  5571 &  1878 & 14156\\
\hline
{\textbf{AddCharacterPersona}} &
    input & 1419 & 20093747 & 11694 & 14090  & 14156\\
  &  labels  &       24.87 &  352124   &   4867   &  1401  & 14156\\
\hline
{\textbf{AddCharacter}} &
    input   &   1262  &   17885361 &  11736   &   14114  &  14170\\
  &  labels  &   9.453  &   133945 & 1568  &   2517  &  14170\\
\hline
{\textbf{AddCharacterWearing}} &
    input &  1651  &  3882027  &    11988    & 2351 &    2351\\
  &  labels  &      12.03  &  28292   &    1009   &       1359  &   2351\\
\hline
{\textbf{AddCharacterWielding}} &
    input  &  1661 & 1538167  &   9861  &     926 &   926\\
   & labels  &    12.06   &   11171    &    655  &   613  &  926\\
\hline
{\textbf{AddObjectContains}} &
    input &  1662  &  452005 & 5771  &  272 & 272\\
 &   labels  &   11.64  & 3167  &   254  &   163  & 272\\
\hline
{\textbf{AddObjectDescription}} &
    input &     1425  & 20164728  &  11686  &  14118 & 14154\\
  &  labels & 14.52  & 205505 & 4522 & 2079  & 14154\\
\hline
{\textbf{AddObject}} &
    input  &   1346  &   19062005  & 11717 &  14115 &  14160\\
  &  labels  &  10.48  & 148388 & 1506 &  2499  &  14160\\
\hline
{\textbf{ObjectsAttributes}} &
    input &      1439  &  20382133  &  11728  & 14150 &  14161\\
  &  labels  &  12.21  & 172969  & 1149  &  5224  & 14161\\
\hline
{\textbf{RoomBackstory}} &
input  &    1391 &19681348 &   10968  & 14003 &14152\\
& labels &    46.82 &  662660 &    3611 &    454 &14152\\
\hline
{\textbf{RoomDescription}} &
input &     1388 &19637390 &   10881 &  13820 &14149\\
& labels &    50.16 &  709730 &    3876 &    459 &14149\\
\hline
{\textbf{GameActions}} &
input &     1389 &  1767204 &   10307 &   1272 & 1272\\
& labels &    13.33 &  16952 &    1112 &    549 & 1272\\
\hline
{\textbf{GameActionsNarration}} &
    input & 1376  &  1770583 & 10242 &  1287  & 1287\\
  &  labels & 6.81  &  8765 &  1012  & 1114  &  1287\\
\hline
{\textbf{InvalidSelfPlay}} &
    input  &  1621  & 8486835 & 15209  & 5235 &  5236\\
   &  labels  & 4.893   &  25620 &   438    &  92  &  5236\\
\hline
{\textbf{InvalidSelfPlayNarration}} &
    input  &     1622  &  8491983 & 15208   &   5235  &  5236\\
   &  labels  &    13.09   & 68523  &  1711     &  4177 &  5236\\
\hline
{\textbf{SelfPlayActions}} &
input &     1848 & 104756348 &   15208 & 56633 &56683 \\
& labels &    15.31 &  867942 &    2241 &   8152 &56683   \\ 
\hline
{\textbf{SelfPlayActionsNarration}} &
input &     1849 & 104812888 &  15207 &  56633 &56683 \\
& labels &    7.251 &  411004 &    2058 &  22569 &56683 \\
\hline
{\textbf{UseEventActions}} &
input &     1565 &13322267 &   17398 &   8512 & 8512\\
& labels &    185.8 & 1581621 &   13798 &   7060 & 8512\\
\hline
{\textbf{UseEventActionsNarration}} &
input  &    1570 & 13360750  &  17385  & 8512  & 8512\\
& labels  &  24.83   & 211317  & 9571 &  8512  & 8512\\
\hline
{\textbf{ALL}} &

input &    1614 &426763460 &   22776 & 263514  &    264333\\
& labels &  22.35 &  5907548 &   19420 &  72418   &   264333\\
\end{tabular}
\caption{Dataset Statistics for all of our constructed tasks. Values are reported for the training splits in each case.}
\label{tab:dataset_stats}
\end{table*}

\subsection{Model Setup}

We perform experiments on our task setups using BART \citep{Lewis2020BARTDS} as our language model architecture. 

We find that our results are sensitive to the relative weighting of our variety of tasks, but are largely very capable of learning our graph-text representation and making cohesive predictions using its format. 
As every type of edge in the $Graph$ can be dropped out, initial ablations to provide an experimental setting led us to the dropout configuration described in Appendix Table \ref{tab:dropout-config}. After freezing these options for training, we consider three sets of input tasks:
\begin{itemize}
\itemsep0em 
    \item \textbf{Narrations Only:} using Narrations Tasks for both LIGHT game events and \textsc{UseEvents}.
    \item \textbf{Narrations and Graph Updates:} using Action Tasks for both LIGHT game events and \textsc{UseEvents} and to predict both Narrations and Graph Updates.
    \item \textbf{Narrations, Updates, and Environment:} using all of the tasks, thus providing additional grounding with the Environment Tasks.
\end{itemize}
We also test three options for dropping the $Graph$ from input context: removing it entirely in 25\%, 50\% or 100\% of examples. We do not consider a dropout of 0\% as we would like to test if learning the $Graph$ improves performance even when the $Graph$ is not provided in the context. 

The combination of Narrations Only and a $Graph$ dropout of 100\% represents our baseline, a model trained to predict narrations that has never seen the direct graph context.

\section{Evaluation and Results}

A core research question in this work is to evaluate if models trained with $Graph$ context and on graph-grounding tasks perform better than language models that are not trained on either $Graph$ inputs or targets, and whether or not those trends remain even without having $Graph$ access in the context at inference time.

\subsection{Automatic Metrics}

\begin{table*}[!ht]
\small 
\centering
\begin{tabular}{l | l  l | r | r | r | r || r | r}
\rule{0pt}{50pt}
id & \multicolumn{1}{l}{Training Task} & 
Graph Dropout &
\rot{Game Action \\~~~Narration} & \rot{Game Action \\~~~Graph Update} & \rot{\textsc{UseEvents} \\~~~Narration} & \rot{\textsc{UseEvents} \\~~~Graph Update} & \rot{$Graph$-less Game \\~~~Action Narration} & \rot{$Graph$-less \textsc{UseEvents} \\~~~Narration}\\
\hline
A & Narrations &  100\%  & 1.346 & - & 6.028 & - & 1.282 & 6.297 \\
B & Only       &  50\% & 1.407 & - & 6.043 & - & 1.265 & 6.351 \\
C &            & 25\% & 1.226 & - & 5.967 & - & 1.250 & 6.360 \\
\hline
D & Narrations + & 100\%  & 1.305 & 1.415 & 6.014 & 2.486 & 1.279 & 6.182 \\
E &  Graph Updates  & 50\%  & 1.293 & 1.234 & \textbf{5.771} & 2.243 & 1.258 & \textbf{6.060} \\
F &   & 25\% & 1.315 & 1.221 & 5.789 & 2.202 & \textbf{1.237} & 6.251 \\
\hline
G & Narrations + & 100\%  & 1.380 & 1.216 & 6.400 & 2.322 & 1.351 & 6.609 \\
H & Graph Updates +  & 50\%  & 1.289 & 1.189 & 5.908 & 2.139 & 1.261 & 6.170 \\
I &  Environment     & 25\% & \textbf{1.195} & \textbf{1.171} & 5.839 & \textbf{2.025} & 1.264 & 6.274 \\
\hline
\end{tabular}
\caption{Automatic Evaluation Results: We report the perplexity of models evaluated on Narrations and Graph Updates for standard Game Actions as well as \textsc{UseEvents}, alongside predicting Narrations without any graph content (``Graph-less'') at evaluation time.}
\label{tab:metrics}
\end{table*}

We first measure perplexity
over four tasks and two ablations: Game Action Narrations, Game Action Updates, \textsc{UseEvent} Narrations, \textsc{UseEvent} Graph Updates, and versions of the two narrations tasks with the $Graph$ context entirely omitted. Model \textbf{A} serves as our baseline, and the graphless settings are our core tests. Note that outside the Graphless cases, models are tested with the same graph dropout setting as they are trained on.

Results are reported in Table \ref{tab:metrics}. 
The best performing model across many automatic metrics is Model \textbf{I}, with best performance across all Graph Updates Tasks. Models \textbf{E} and \textbf{F} perform best in the graphless tests of \textsc{UseEvents} and Game Actions respectively. 
Thus, we find that training with graph knowledge generally provides better performance, even in the case where this knowledge is not given at inference time.

We generally see good perplexity performance on our Game Action Tasks, with both narrations and Graph Updates at low values. \textsc{UseEvent} narrations have a significantly higher perplexity than the more formulaic Game Action narrations, and \textsc{UseEvent} graph updates have a notably higher perplexity than game actions. 

This follows expectations, as all Game Actions are themselves built from a formulaic set of templates, and should be somewhat easily learnable. \textsc{UseEvent} graph updates are more  diverse and much longer than Game Actions graph updates, but they are built from the same dictionary of triples. \textsc{UseEvent} narrations, in comparison, are the most varied task and have the highest perplexity values.

\subsection{Human Evaluation}
\label{sec:human-eval}

\begin{table*}[!ht]
\small 
\centering
\begin{tabular}{l | r | r | r | r | r || r | r | r | r}
\multicolumn{6}{c||}{Human Evaluations} & \multicolumn{4}{c}{
Automatic Evaluation of Narrations} \\
& \multicolumn{1}{c|}{Training} & \multicolumn{1}{c|}{Graph} & \multicolumn{2}{c|}{Inconsistent} & \multicolumn{1}{c||}{All} & \multicolumn{2}{c|}{With $Graph$} & \multicolumn{2}{c}{Without $Graph$}\\
id & \multicolumn{1}{c|}{Task} & Dropout & Action & Setting & Good & PPL & F1 & PPL & F1 \\
\hline
I & All & 25\% & \textbf{0.04} & \textbf{0.23} & \textbf{0.72} & 1.107 & 0.934 & 1.261 & 0.612  \\
E & Narration + Graph Updates & 50\% & 0.18 & 0.33 & 0.57 & 1.148  & 0.897 & 1.258 & 0.639 \\
A & Narration Only & 100\% & 0.20 & \textbf{0.23} & 0.57 & - & - & 1.282 & 0.666\\
H & All & 50\% & 0.25 & 0.24 & 0.51 & 1.151 & 0.910 & 1.261 & 0.660\\
F & Narration  +  Graph Updates & 25\% & 0.06 & 0.53 & 0.44 & 1.170  & 0.847 & 1.237 & 0.675 \\
\hline
\end{tabular}
\caption{Human Evaluation Results: We evaluate a subset of the models in the setting without graph information given in the context, 100 interactions each. We report the fraction of evaluators that marked responses as inconsistent with the setting, inconsistent with the action, both, or good. We also include comparison automatic metrics for our Graph Actions Narrations task with and without $Graph$ access.}
\label{tab:h_evals}
\end{table*}

We collected human evaluations of our best performing models by asking crowdworkers to take a single turn in LIGHT using the model as a game engine, where they then annotate the predicted narrations for consistency with the environment and with the entered action. In this case being inconsistent {\em with the action} means that the returned narration's outcome does not follow from the requested action, for instance ``chop tree with the axe'' returning responses like ``You pick up the axe" or ``You eat the apple''. Being inconsistent {\em with the setting} means the returned narration's outcome does not follow from what one expects of the setting, for instance ``eat the apple'' returning ``You can't eat that!'' or ``You don't have an apple to eat'' when the context notes the character is holding an apple. Workers were able to mark each turn with multiple problem labels. The task interface is shown in Figure \ref{fig:human_eval}.

For this stage we chose Model \textbf{A} to be the baseline, and models \textbf{I}, \textbf{E}, and \textbf{F} as top performing models that were trained with graph grounding tasks. We additionally test model \textbf{H}, which was our highest performing model on Game Action Narrations without $Graph$ in context of our models trained on all tasks.

Results are provided in Table \ref{tab:h_evals}. We report the proportion of examples marked with two core failure modes, inconsistent action and inconsistent setting, alongside the proportion of events with no issues marked. We additionally provide additional automatic metrics for these models, both perplexity and F1 overlap for Game Action Narrations where the $Graph$ context is never dropped out and always dropped out, for the purpose of comparison.

Model \textbf{I}, a model trained on all of our grounding tasks, notably outperforms all of the other models on All Good and the other human evaluation metrics, and also interestingly has the best performance on all Game Action Narration with $Graph$ context metrics.

\subsection{Analysis}

\subsubsection{Human Evaluation Behaviors}

There were a few notable evaluator behaviors during the human evaluations that make it different from the training setup. The first one is that over 80\% of the test interactions workers provided were `single-target', using one element in their target. These events, like ``knock on door'', are entirely unlike the \textsc{UseEvents} data. Roughly half of these were directly verbs already covered in the LIGHT game engine, while the rest were either synonym verbs (like `swing at tiger' instead of `hit tiger') or entirely new verbs (like a spider that asked to `climb the wall'). Secondly, nearly all (>95\%) of the interactions evaluators used would be expected to occur successfully in the setting, whereas our training set had a higher proportion of invalid actions. 

From this perspective, we suspect that the Game Action automatic metrics are a closer indicator of a model's potential success on our human evaluation task than the \textsc{UseEvent} metrics.

\subsubsection{Human Evaluation Performance}

\begin{figure*}[!ht]
    \centering
    \includegraphics[width=\textwidth]{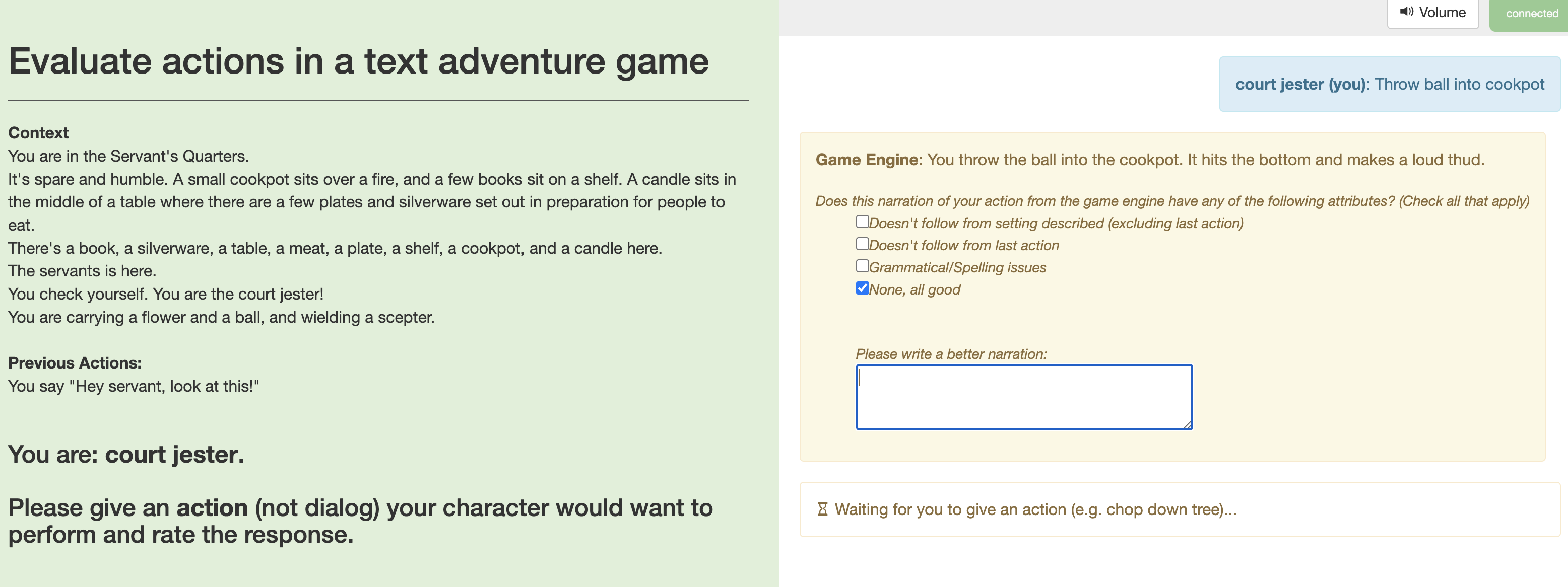}
    \caption{Human Evaluation Task for Narration Predictions shown to crowdworkers.}
    \label{fig:human_eval}
\end{figure*}

The best performing model in human evaluations was Model \textbf{I}, which was trained on all training tasks. The additional performance gains are primarily driven by a reduction in inconsistent action, which implies that the model responses better followed the objects and outcomes as requested in the action. Being inconsistent with the setting remains level, however it is possible that the model improved across examples that would have been labeled with inconsistent setting in the baseline, but is now making setting mistakes for newly understood actions. 

Model \textbf{I} was also the \textit{only} model that significantly outperformed the baseline Model \textbf{A} in human evaluations. Note though that the human evaluation setting was nearly equivalent to the training setting for Model \textbf{A}, and as such Model \textbf{I}'s performance implies that our introduced training tasks and $Graph$ context can help performance here.

There is a visible trend between the Game Action Narration with $Graph$ context perplexity (and to a lesser degree F1) and a model's all good proportion. Models that seem to best grasp how to use the $Graph$ context to produce a narration end up also producing good narrations even when the graph is not present. This is in contrast to the automatic metrics for Game Action Narration without $Graph$ context, which seem to have a slight inverse relationship.

Given the gap in automatic metrics between the $Graph$-grounded and $Graph$-less settings, an additional comparison could also be collected to see how the $Graph$-trained models perform on human evaluations if provided the $Graph$ context, however this would be future work.

\subsubsection{Observations on Model Behaviors}

While analyzing the human evaluation results, we noticed a few differing trends in each model's responses. Each of the following trends are in relation to the behavior of Model \textbf{I}, the model trained on all tasks with low $Graph$ dropout.

\textbf{Model E}, trained on Narration and Update tasks with moderate $Graph$ dropout: This model appears to always treat \textsc{UseEvent}-style interactions as being invalid. For example, \textit{``You cannot cut the map with the knife, the knife is too soft and it would just break''}. Given that invalid object-object interactions have a more formulaic pattern than the many success cases, these are likely the easier ones to learn. This behavior may explain being the best at \textsc{UseEvent} Narration perplexity. These also ended up being the answers most frequently marked as inconsistent with both the action and the setting.

\textbf{Model A}, the baseline: also learned to make \textsc{UseEvent}-style interactions unsuccessful, however it seemed to have a harder time understanding the multi-object nature of the events. Instead, it would fail the action and often with an incorrect interpretation of the action. For example, \textit{``You can't find a 'knife' here that you can cut''}.

\textbf{Model F}, trained on Narration and Update tasks with low $Graph$ dropout: This was all-around pretty unsuccessful at handling interactions outside of the basic LIGHT Game Actions. For any interaction outside that set, it would respond with ``You don't have <object> that you can <verb>''. It also had some unexpected additional disfluencies, for instance once claiming \textit{``You can't chop down a tree! It's not a tree! It's a tree! It's not even a branch! It's a tree!''}.

\textbf{Model H}, trained on all tasks with moderate $Graph$ dropout: appears to act like something between Models \textbf{E} and \textbf{F}, marking \textsc{UseEvent}-style actions as invalid less frequently than the former, and was able to react to new actions more frequently than the latter.

\section{Conclusions and Future Work}

This work has studied the use of graph knowledge when training commonsense world models. We have found that adding in $Graph$ context during training time \textit{can} help language models produce quality narrations \textit{without} the graph being given at evaluation time. This is observed in both automatic metrics and human evaluations.

The datasets we have built are challenging as can be seen from the reported evaluation metrics, in particular the ``all good'' proportion over a user's expected set of interactions. We will publicly release these tasks for research by the community.
Still, future work could explore further extending the action space of our datasets and collecting a larger set of rich action narrations, as well as evaluating multiple turns of narration. Further, human evaluations could be developed that more accurately capture long-tail interactions such as those in the \textsc{UseEvents} set.

Future modeling work could focus on seeing which problems are resolved by scaling the models, or if there is a need to have models explicitly create an internal representation for the graph state.

\section{Ethical Considerations}
As we collected \textsc{UseEvents} on top of the underlying LIGHT dataset \cite{urbanekLearningSpeakAct2019}, we have to consider that this dataset and its tasks inherit and may extend some of the underlying issues in LIGHT, for example in terms of potential toxicity or bias in characters, dialogue, or actions and events. Work has already been done to address some of these issues particularly when it comes to character representation and dialogue \citep{dinan2019queens}, however no analysis has been done on the underlying objects or learned relationships. Further, our current tasks are learning from the dialogues in the original LIGHT paper, for which the settings do not include the noted mitigations.

We note however that the \textsc{UseEvents} dataset does not generally refer to people, and is thus less likely to show these types of biases, but they could still be present in certain cases. Further, our Action Tasks could be set to use any of the LIGHT dialogue datasets (including ones with debiasing) for use in a live deployment or release.

\section*{Acknowledgements}
Thanks to Justin Pinero for building out the initial crowdsourcing interfaces. Thanks to the ParlAI framework \citep{parlai},  which made it easy to implement the tasks for this work. 

\bibliographystyle{acl_natbib}
\bibliography{references.bib}

\begin{thebibliography}{17}
\expandafter\ifx\csname natexlab\endcsname\relax\def\natexlab#1{#1}\fi

\bibitem[{Ammanabrolu et~al.(2019)Ammanabrolu, Broniec, Mueller, Paul, and
  Riedl}]{ammanabrolu-etal-2019-toward}
Prithviraj Ammanabrolu, William Broniec, Alex Mueller, Jeremy Paul, and Mark
  Riedl. 2019.
\newblock \href {https://aclanthology.org/2019.ccnlg-1.1} {Toward automated
  quest generation in text-adventure games}.
\newblock In \emph{Proceedings of the 4th Workshop on Computational Creativity
  in Language Generation}, pages 1--12, Tokyo, Japan. Association for
  Computational Linguistics.

\bibitem[{Ammanabrolu and Riedl(2021)}]{ammanabroluModelingWorldsText2021}
Prithviraj Ammanabrolu and Mark~O. Riedl. 2021.
\newblock Modeling {{Worlds}} in {{Text}}.
\newblock \emph{NeurIPS Datasets and Benchmarks}.

\bibitem[{C{\^o}t{\'e} et~al.(2018)C{\^o}t{\'e}, K{\'a}d{\'a}r, Yuan, Kybartas,
  Barnes, Fine, Moore, Hausknecht, Asri, Adada, Tay, and
  Trischler}]{coteTextWorldLearningEnvironment2018}
Marc-Alexandre C{\^o}t{\'e}, {\'A}kos K{\'a}d{\'a}r, Xingdi Yuan, Ben~A.
  Kybartas, Tavian Barnes, Emery Fine, James Moore, M.~Hausknecht, Layla~El
  Asri, Mahmoud Adada, Wendy Tay, and Adam Trischler. 2018.
\newblock \href {https://doi.org/10.1007/978-3-030-24337-1_3} {{{TextWorld}}:
  {{A Learning Environment}} for {{Text-based Games}}}.
\newblock In \emph{{{CGW}}@{{IJCAI}}}.

\bibitem[{Dinan et~al.(2019)Dinan, Fan, Williams, Urbanek, Kiela, and
  Weston}]{dinan2019queens}
Emily Dinan, Angela Fan, Adina Williams, Jack Urbanek, Douwe Kiela, and Jason
  Weston. 2019.
\newblock Queens are powerful too: Mitigating gender bias in dialogue
  generation.
\newblock \emph{arXiv preprint arXiv:1911.03842}.

\bibitem[{{Latitude AI}(2019)}]{aidungeon}
{Latitude AI}. 2019.
\newblock {AI Dungeon 2}.
\newblock \url{https://aidungeon.io/}.
\newblock Accessed: 2022-12-29.

\bibitem[{Lewis et~al.(2020)Lewis, Liu, Goyal, Ghazvininejad, Mohamed, Levy,
  Stoyanov, and Zettlemoyer}]{Lewis2020BARTDS}
Mike Lewis, Yinhan Liu, Naman Goyal, Marjan Ghazvininejad, Abdelrahman Mohamed,
  Omer Levy, Veselin Stoyanov, and Luke Zettlemoyer. 2020.
\newblock Bart: Denoising sequence-to-sequence pre-training for natural
  language generation, translation, and comprehension.
\newblock In \emph{Annual Meeting of the Association for Computational
  Linguistics}.

\bibitem[{Lin et~al.(2019)Lin, Shen, Xing, Zhou, and Ren}]{Lin2019CommonGenAC}
Bill~Yuchen Lin, Minghan Shen, Yu~Xing, Pei Zhou, and Xiang Ren. 2019.
\newblock Commongen: A constrained text generation dataset towards generative
  commonsense reasoning.
\newblock \emph{ArXiv}, abs/1911.03705.

\bibitem[{Liu et~al.(2021)Liu, Wan, He, Peng, and
  Yu}]{liuKGBARTKnowledgeGraphAugmented2021}
Ye~Liu, Yao Wan, Lifang He, Hao Peng, and Philip~S. Yu. 2021.
\newblock {{KG-BART}}: {{Knowledge Graph-Augmented BART}} for {{Generative
  Commonsense Reasoning}}.
\newblock \emph{Proceedings of the AAAI Conference on Artificial Intelligence},
  35(7):6418--6425.

\bibitem[{Miller et~al.(2017)Miller, Feng, Fisch, Lu, Batra, Bordes, Parikh,
  and Weston}]{parlai}
Alexander~H Miller, Will Feng, Adam Fisch, Jiasen Lu, Dhruv Batra, Antoine
  Bordes, Devi Parikh, and Jason Weston. 2017.
\newblock Parlai: A dialog research software platform.
\newblock \emph{arXiv preprint arXiv:1705.06476}.

\bibitem[{Prabhumoye et~al.(2020)Prabhumoye, Li, Urbanek, Dinan, Kiela, Weston,
  and Szlam}]{Prabhumoye2020ILY}
Shrimai Prabhumoye, Margaret Li, Jack Urbanek, Emily Dinan, Douwe Kiela, Jason
  Weston, and Arthur~D. Szlam. 2020.
\newblock I love your chain mail! making knights smile in a fantasy game world:
  Open-domain goal-oriented dialogue agents.
\newblock \emph{ArXiv}, abs/2002.02878.

\bibitem[{Sap et~al.(2019)Sap, Le~Bras, Allaway, Bhagavatula, Lourie, Rashkin,
  Roof, Smith, and Choi}]{sapATOMICAtlasMachine2019}
Maarten Sap, Ronan Le~Bras, Emily Allaway, Chandra Bhagavatula, Nicholas
  Lourie, Hannah Rashkin, Brendan Roof, Noah~A. Smith, and Yejin Choi. 2019.
\newblock \href {https://doi.org/10.1609/aaai.v33i01.33013027} {{{ATOMIC}}: An
  atlas of machine commonsense for if-then reasoning}.
\newblock In \emph{Proceedings of the {{Thirty-Third AAAI Conference}} on
  {{Artificial Intelligence}} and {{Thirty-First Innovative Applications}} of
  {{Artificial Intelligence Conference}} and {{Ninth AAAI Symposium}} on
  {{Educational Advances}} in {{Artificial Intelligence}}},
  {{AAAI}}'19/{{IAAI}}'19/{{EAAI}}'19, pages 3027--3035, {Honolulu, Hawaii,
  USA}. {AAAI Press}.

\bibitem[{Sutskever et~al.(2014)Sutskever, Vinyals, and Le}]{seq2seq}
Ilya Sutskever, Oriol Vinyals, and Quoc~V Le. 2014.
\newblock Sequence to sequence learning with neural networks.
\newblock \emph{Advances in neural information processing systems}, 27.

\bibitem[{Urbanek et~al.(2019)Urbanek, Fan, Karamcheti, Jain, Humeau, Dinan,
  Rockt{\"a}schel, Kiela, Szlam, and Weston}]{urbanekLearningSpeakAct2019}
Jack Urbanek, Angela Fan, Siddharth Karamcheti, Saachi Jain, Samuel Humeau,
  Emily Dinan, Tim Rockt{\"a}schel, Douwe Kiela, Arthur Szlam, and Jason
  Weston. 2019.
\newblock \href {https://doi.org/10.18653/v1/D19-1062} {Learning to {{Speak}}
  and {{Act}} in a {{Fantasy Text Adventure Game}}}.
\newblock In \emph{Proceedings of the 2019 {{Conference}} on {{Empirical
  Methods}} in {{Natural Language Processing}} and the 9th {{International
  Joint Conference}} on {{Natural Language Processing}} ({{EMNLP-IJCNLP}})},
  pages 673--683, {Hong Kong, China}. {Association for Computational
  Linguistics}.

\bibitem[{Urbanek and Ringshia(2023)}]{mephisto}
Jack Urbanek and Pratik Ringshia. 2023.
\newblock \href {https://doi.org/10.48550/ARXIV.2301.05154} {Mephisto: A
  framework for portable, reproducible, and iterative crowdsourcing}.

\bibitem[{Zellers et~al.(2018)Zellers, Bisk, Schwartz, and
  Choi}]{Zellers2018SWAGAL}
Rowan Zellers, Yonatan Bisk, Roy Schwartz, and Yejin Choi. 2018.
\newblock Swag: A large-scale adversarial dataset for grounded commonsense
  inference.
\newblock In \emph{Conference on Empirical Methods in Natural Language
  Processing}.

\bibitem[{Zellers et~al.(2019)Zellers, Holtzman, Bisk, Farhadi, and
  Choi}]{Zellers2019HellaSwagCA}
Rowan Zellers, Ari Holtzman, Yonatan Bisk, Ali Farhadi, and Yejin Choi. 2019.
\newblock Hellaswag: Can a machine really finish your sentence?
\newblock In \emph{Annual Meeting of the Association for Computational
  Linguistics}.

\bibitem[{Zellers et~al.(2021)Zellers, Holtzman, Peters, Mottaghi, Kembhavi,
  Farhadi, and Choi}]{zellersPIGLeTLanguageGrounding2021}
Rowan Zellers, Ari Holtzman, Matthew Peters, Roozbeh Mottaghi, Aniruddha
  Kembhavi, Ali Farhadi, and Yejin Choi. 2021.
\newblock \href {https://doi.org/10.18653/v1/2021.acl-long.159} {{{PIGLeT}}:
  {{Language Grounding Through Neuro-Symbolic Interaction}} in a {{3D World}}}.
\newblock In \emph{Proceedings of the 59th {{Annual Meeting}} of the
  {{Association}} for {{Computational Linguistics}} and the 11th
  {{International Joint Conference}} on {{Natural Language Processing}}
  ({{Volume}} 1: {{Long Papers}})}, pages 2040--2050, {Online}. {Association
  for Computational Linguistics}.

\end{thebibliography}

\appendix

\section{Graph Dataset Format}
\label{sec:graph-dataset-format}

\begin{figure*}[ht]
    \centering
    \includegraphics[width=\textwidth]{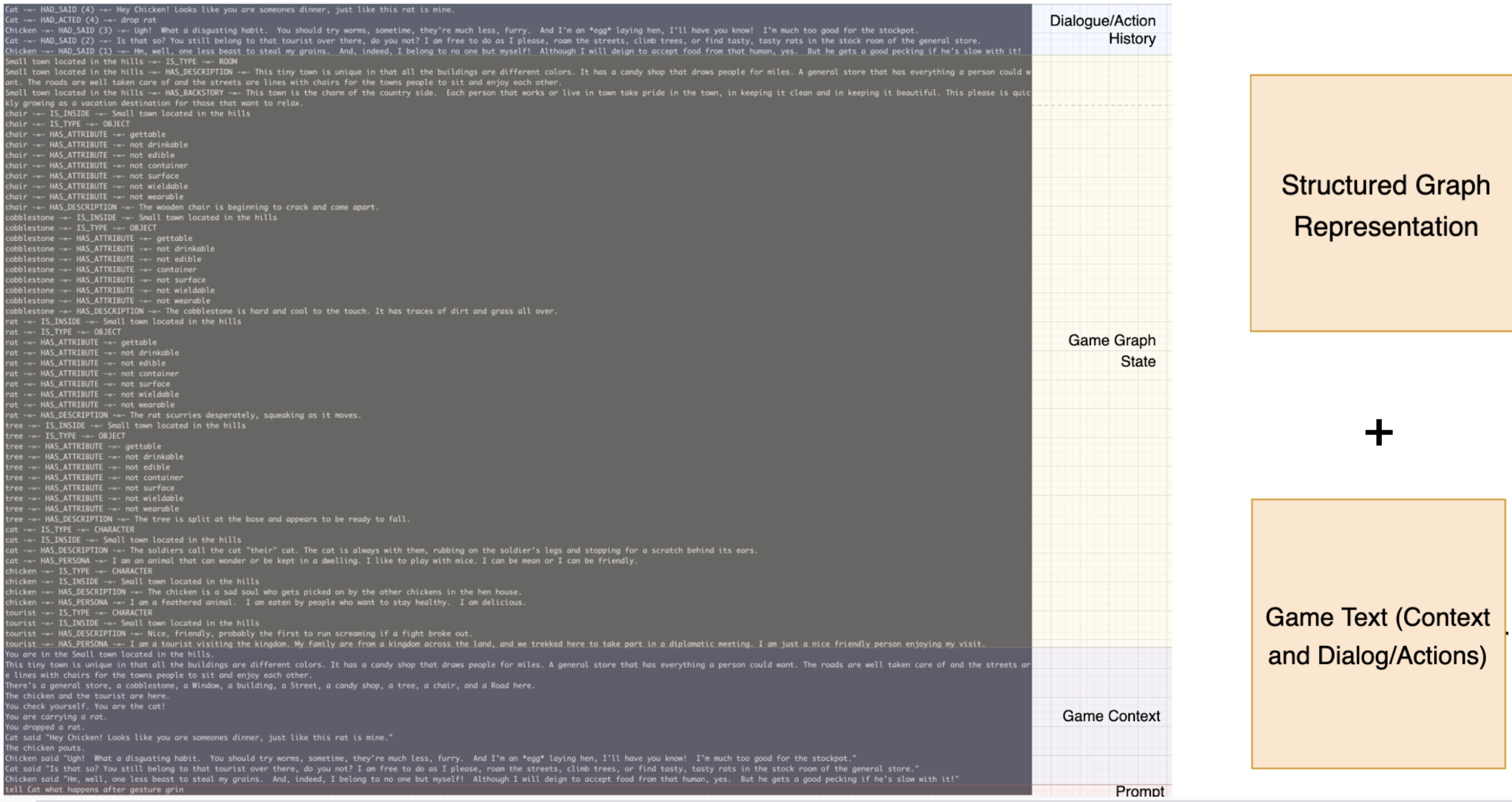}
    \caption{Rough overview of the structure of episodes in the dataset}
    \label{fig:graph_breakdown}
\end{figure*}

We share an overview of the graph dataset format in \ref{fig:graph_breakdown}. The context in these examples is extremely large, and consists primarily of the triples that make up the structured graph.

The complete list of edges is: \textit{IS\_TYPE, IS\_INSIDE, IS\_CARRYING, IS\_WIELDING, IS\_WEARING, HAS\_BACKSTORY, HAS\_PERSONA, HAS\_DESCRIPTION, IS\_DEAD, HAS\_DAMAGE\_LEVEL, HAS\_HEALTH\_LEVEL, HAS\_STRENGTH\_LEVEL, HAS\_PLAYER\_CONTEXT, HAS\_ATTRIBUTE, IS\_GETTABLE, IS\_DRINK, IS\_FOOD, IS\_CONTAINER, IS\_SURFACE, IS\_WEARABLE, IS\_WIELDABLE, HAD\_SAID, HAD\_ACTED, OBSERVED, CONTAINS, CURRENT\_PLAYER}.

The options for graph update label are \textit{ADD: <triple>, DEL: <triple>, NO\_MUTATION}

The list of prompts is: 
\begin{itemize}[noitemsep,topsep=0pt,leftmargin=*]
    \item \textit{narrate from \{observer\} perspective: \{actor\} \{act\}} for graph action narrations
    \item \textit{modify graph after: \{actor\} \{act\}} for graph updates
    \item \textit{describe the room} for room descriptions
    \item \textit{background, describe the room backstory, room backstory} for room backstories
    \item \textit{describe \{name\}, examine \{name\}} for adding physical descriptions
    \item \textit{what is the persona of \{name\}, describe the persona of \{name\}} for adding a persona
    \item \textit{add object, add a new object, suggest a new object} for adding objects
    \item \textit{add object contained by \{name\}, suggest a new object contained by \{name\}} for adding contained objects
    \item \textit{add object wielded by \{name\}, suggest a new object wielded by \{name\}} for adding a wielded object for a character
    \item \textit{add object carried by \{name\}, suggest a new object carried by \{name\}} for adding a carried object for a character
    \item \textit{add object worn by \{name\}, suggest a new object worn by \{name\}} for adding a worn object to a character
    \item \textit{add character, add a new character, suggest a new character} for adding a new character
    \item \textit{what is the type of \{name\}, what type is \{name\}, what type of item is \{name\}, \{name\} is what type} for adding the type for an element
    \item \textit{Is \{name\} gettable?, Can I pick up \{name\}?} for determining if an object is gettable
    \item \textit{Is \{name\} drinkable?, Can I drink \{name\}?} for determining if an object is drinkable
    \item \textit{Is \{name\} edible?, Can I eat \{name\}?, Is \{name\} a food?} for determining if an object is edible
    \item \textit{Is \{name\} container?, Can I put something inside \{name\}?} for determining if an object is a container
    \item \textit{Does \{name\} have usable surface?, Can I put something on \{name\}?} for determining if an object is a surface
    \item \textit{Is \{name\} a weapon?, Can I use \{name\} as a weapon?} for determining if an object is a weapon
    \item \textit{Is \{name\} wearable?, Can I wear \{name\}?} for determining if an object is wearable
\end{itemize}

\noindent Given the complexity and length of the graph representation in our dataset, we choose to allow dropout for all of the edges individually (excepting those involved in the labels). This dropout should also help with overfitting on a dataset of our size. The dropout configuration we use for training in this paper is reported in Table \ref{tab:dropout-config}.

A complete set of statistics for all of the tasks based on these datasets can be found in Table \ref{tab:dataset_stats}.

\begin{table*}[t]
\centering
\begin{tabular}{|l | r| l |}
\hline
\textbf{Dropout Type} & \textbf{Amount} & \textbf{Description} \\
\hline
Room Name & 0.1 & The name of the location \\
Room Description & 0.1 & A few sentences describing a location \\
Room Backstory & 0.1 & Additional description for a location, focused on backstory \\
Room Objects & 0.2 & All objects present in a room, applied per-object  \\
Room Characters & 0.2 & All characters present in a room, applied per-character \\
Contained Objects & 0 & Objects that are contained in other objects, applied per-object \\
Worn Objects & 0 & Objects that are worn by present characters, applied per-object \\
Wielded Objects & 0 & Objects wielded by present characters, applied per-object \\
Carried Objects & 0 & Objects carried by present characters, applied per-object \\
Attribute & 0.1 & Any object attributes, such as \texttt{is\_container} \\
Persona & 0.1 & Full personas of present characters \\
Physical Description & 0.1 & Physical descriptions of both characters and objects \\
Character Inside Room & 0.1 & The label of a character being inside of the room \\
Character Type & 0.1 & The label noting that a provided name is a character \\
Object Inside Room & 0.1 & The label of an object being inside of the room \\
Object Type & 0.1 & The label noting that a provided name is an object \\
Dialogue History & 0.25 & The text dialogue preceding the current timestep \\
State Mutations History & 0.25 & Any graph updates that have occurred previously \\
Graph State & 0.25 & All of the tuples that are not dialogue or mutation history \\
Game Text & 0.25 & The full-text context format a user would normally see \\
\hline
\end{tabular}
\caption{Dropout configurations for standard training jobs}
\label{tab:dropout-config}
\end{table*}

\section{LIGHT \textsc{UseEvent} Collection Methodology}
\label{sec:use-event-collection}

Collecting \textsc{UseEvent} data proved to be rather challenging, as we both wanted to have a diverse set of possible actions as well as well annotated groundings for them. It was impossible to collect these at the same time, as workers would be incentivized to create simpler interactions so that there would be less to annotate. 

Due to this, we split the task into two portions, one to create the interaction and another to ground it. 

\subsection{Narration Collection}
\begin{figure*}[ht]
    \centering
    \includegraphics[width=\textwidth]{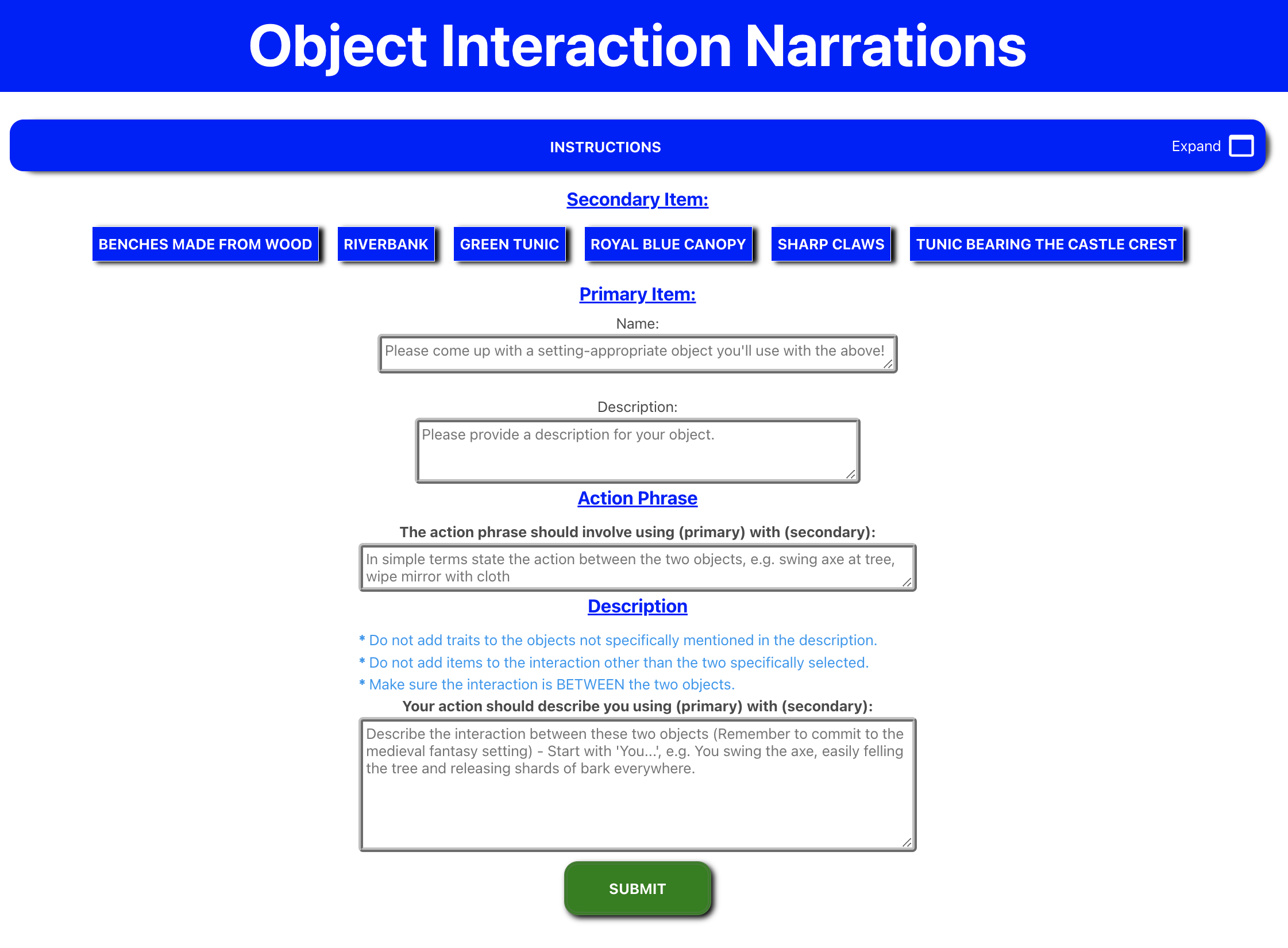}
    \caption{Crowdsourcing task interface for collecting novel object-object interactions}
    \label{fig:collect_narrations}
\end{figure*}

In the first, annotators were provided a list of objects from LIGHT's environment database. They were asked to select one of the objects, and then come up with another object that could be used in an interaction with the first. They would then provide the ``action phrase'' that would trigger this action, and a second-person perspective narration of what they expected would happen. All of this is captured in Figure \ref{fig:collect_narrations}.

\begin{figure*}[ht]
    \centering
    \includegraphics[width=\textwidth]{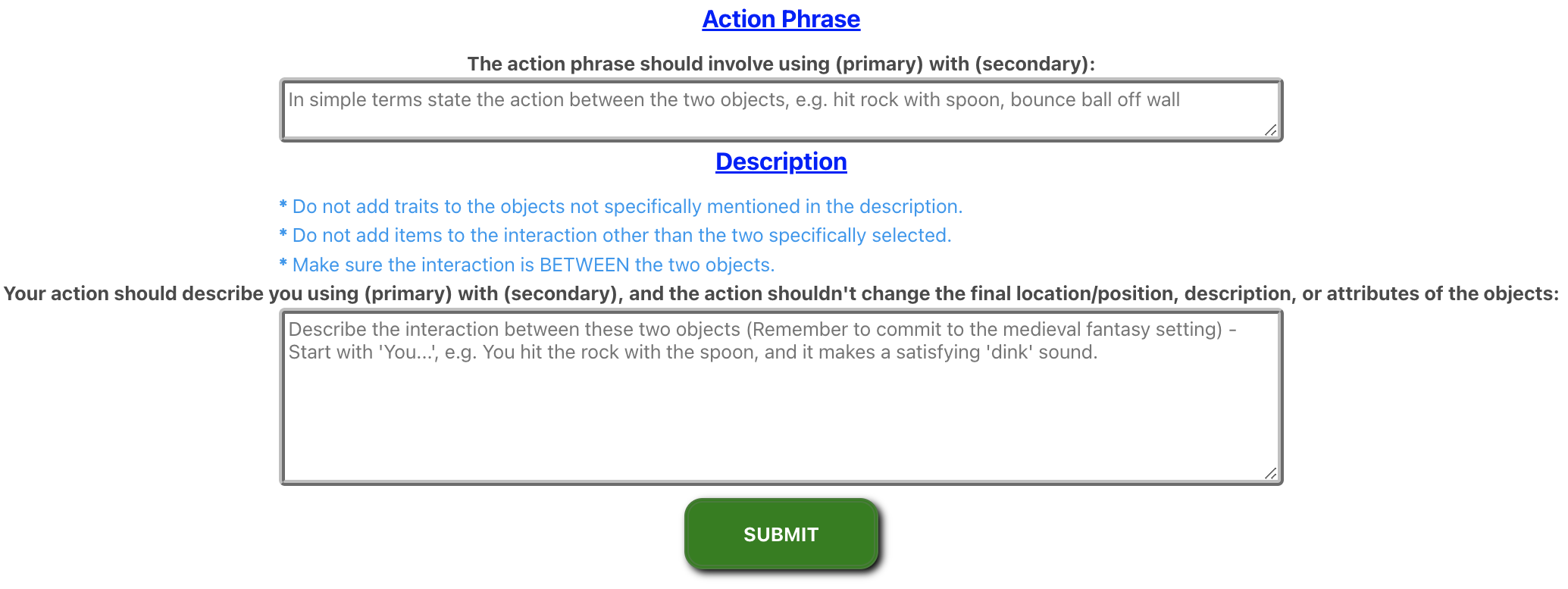}
    \caption{`Boring' variation of Figure \ref{fig:collect_narrations} intended to collect interactions with no updates.}
    \label{fig:collect_boring}
\end{figure*}

\begin{figure*}[ht]
    \centering
    \includegraphics[width=\textwidth]{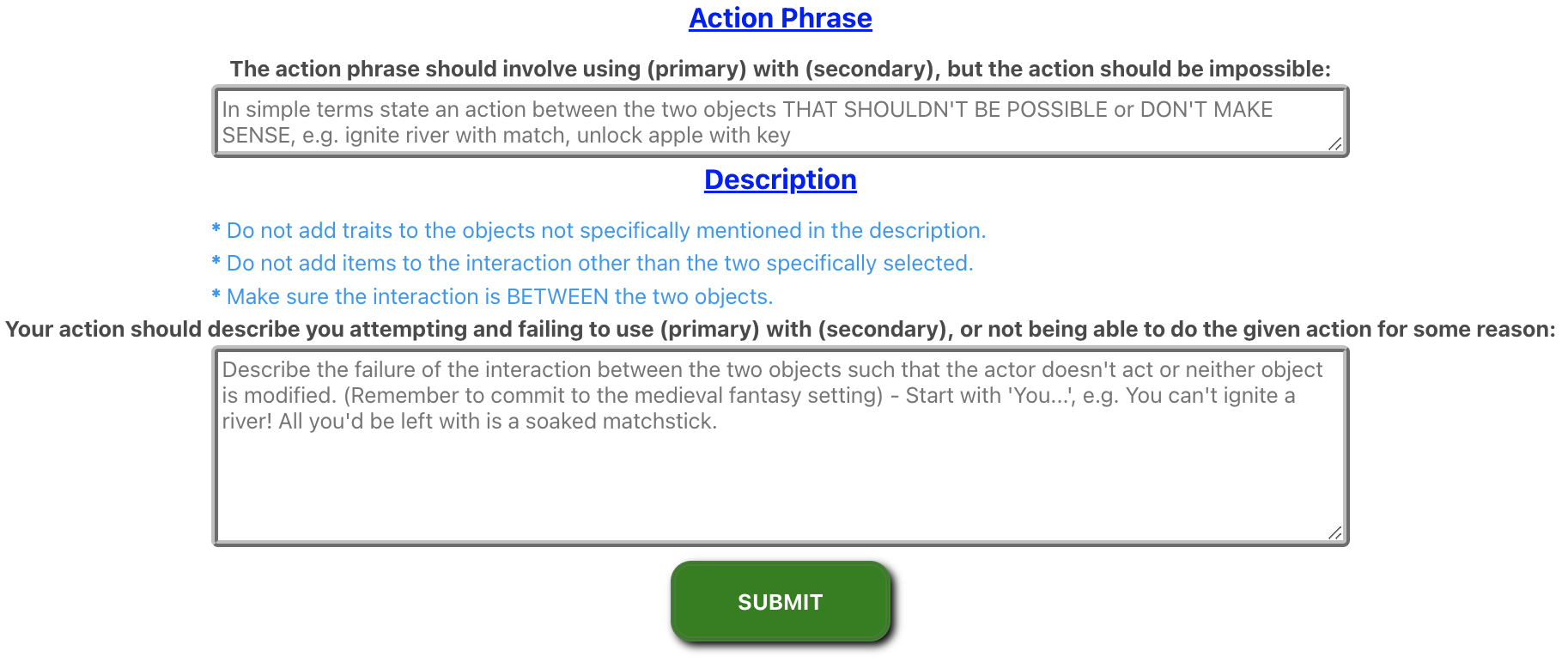}
    \caption{`Failed` variation of Figure \ref{fig:collect_narrations} intended to collect invalid interactions.}
    \label{fig:collect_failed}
\end{figure*}

During collection, we found that the vast majority of interactions collected were successful, in that the desired action was always executed. In order to get some additional diversity in the outcomes for actions, we additionally launched versions of the task that pushed workers to provide ``boring'' or ``failed'' interactions instead. These interface variations are displayed in Figure \ref{fig:collect_boring} and Figure \ref{fig:collect_failed} respectively.

Between the three, we collected roughly 8000 interactions with the first setup, and 2000 total between the other two.

\subsection{Interaction Grounding}

For the grounding step, we found that providing workers with an empty set of fields for the entirety of the grounding resulted in low-quality and short answers. Instead, we prompted a LLM to produce a draft round of values for the majority of the fields required to ground the interaction, then had workers correct them. We filtered out low-quality workers who weren't doing the task as desired, and created an allowlist for those who were completing the task well.

\begin{figure*}[ht]
    \centering
    \includegraphics[width=\textwidth]{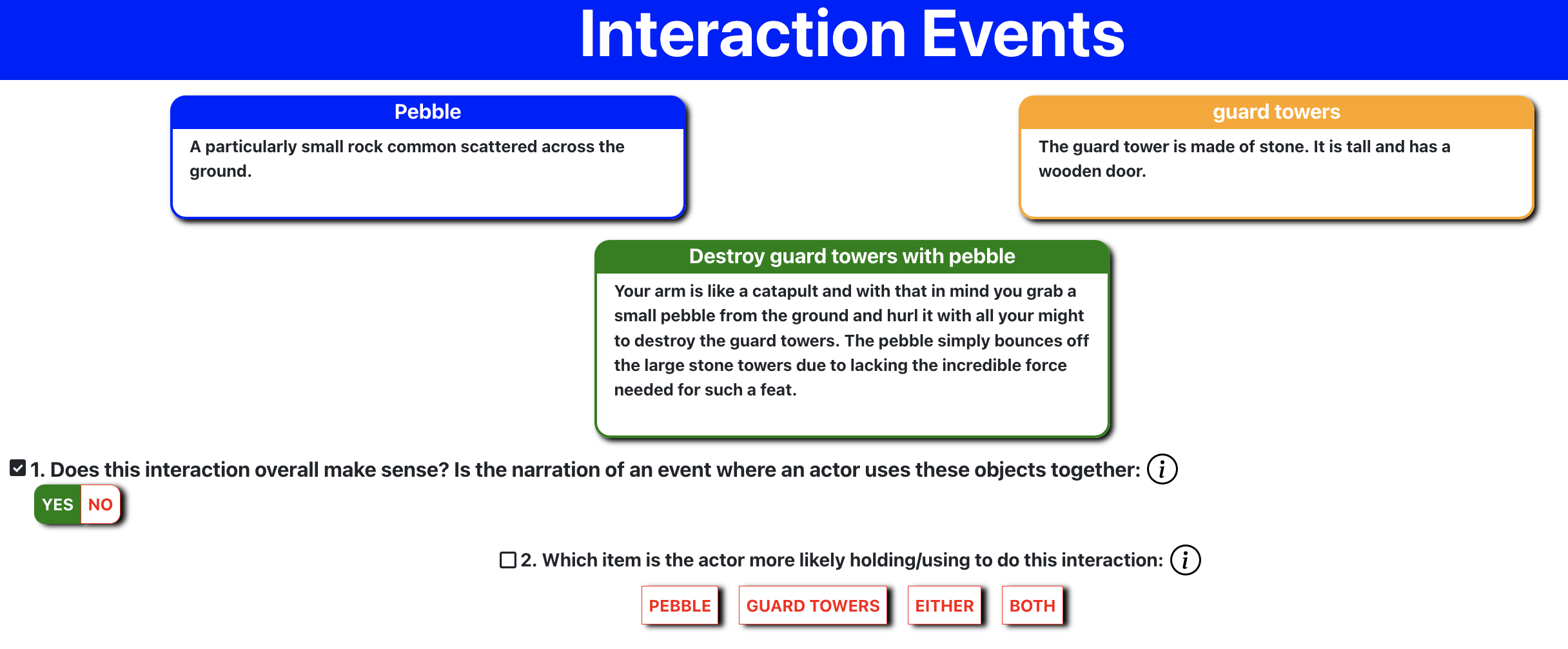}
    \caption{Initial context and validation questions for grounding a collected narration}
    \label{fig:grounding_context}
\end{figure*}

In this phase of the task, workers were first asked to read and understand the initial context, determine if the interaction was valid, and then determine which of the objects an actor would be holding in order to execute the interaction. This is shown in Figure \ref{fig:grounding_context}

\begin{figure*}[ht]
    \centering
    \includegraphics[width=\textwidth]{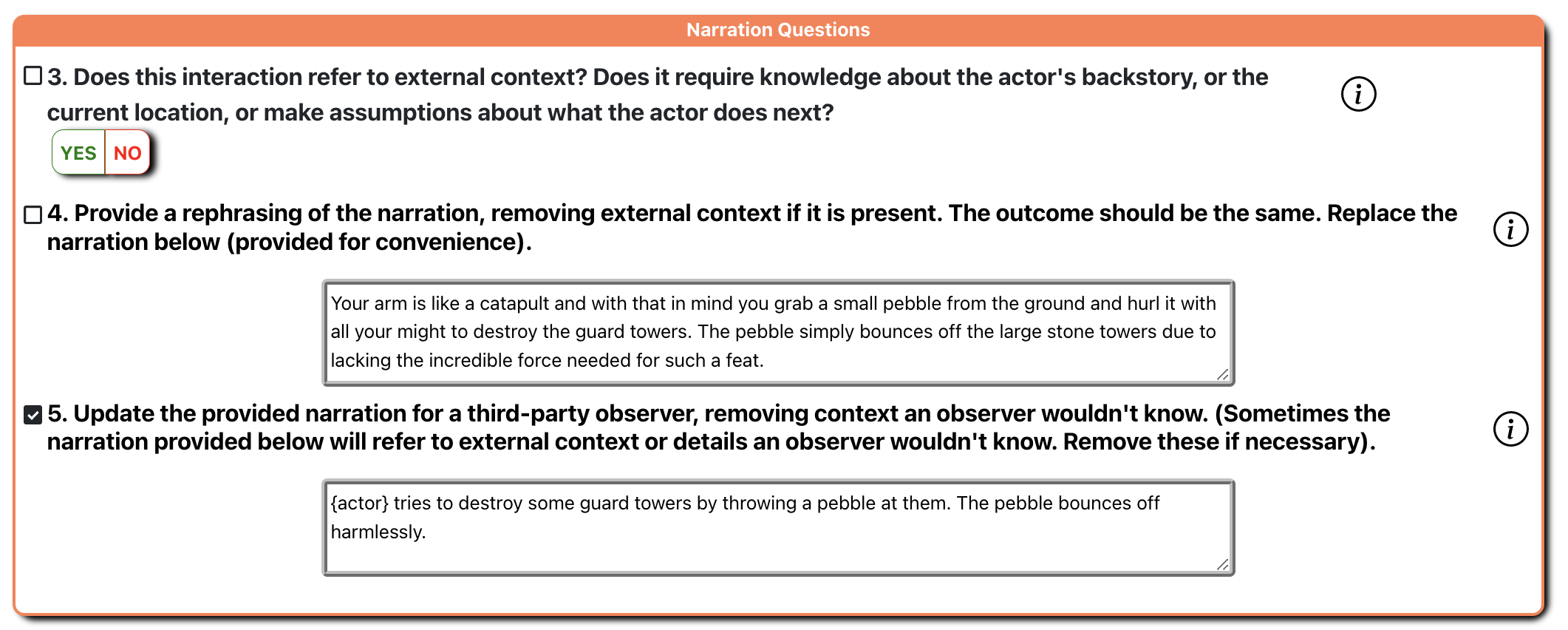}
    \caption{Interface to validate narration and correct automatically generated external narration}
    \label{fig:grounding_narration}
\end{figure*}

They were then asked to determine if the interaction followed our rules in not referring to external context, provide an alternate narration of the event that would have the same outcome, and produce a version of the narration for a third party observing the interaction. This is shown in Figure \ref{fig:grounding_narration}

\begin{figure*}[ht]
    \centering
    \includegraphics[width=\textwidth]{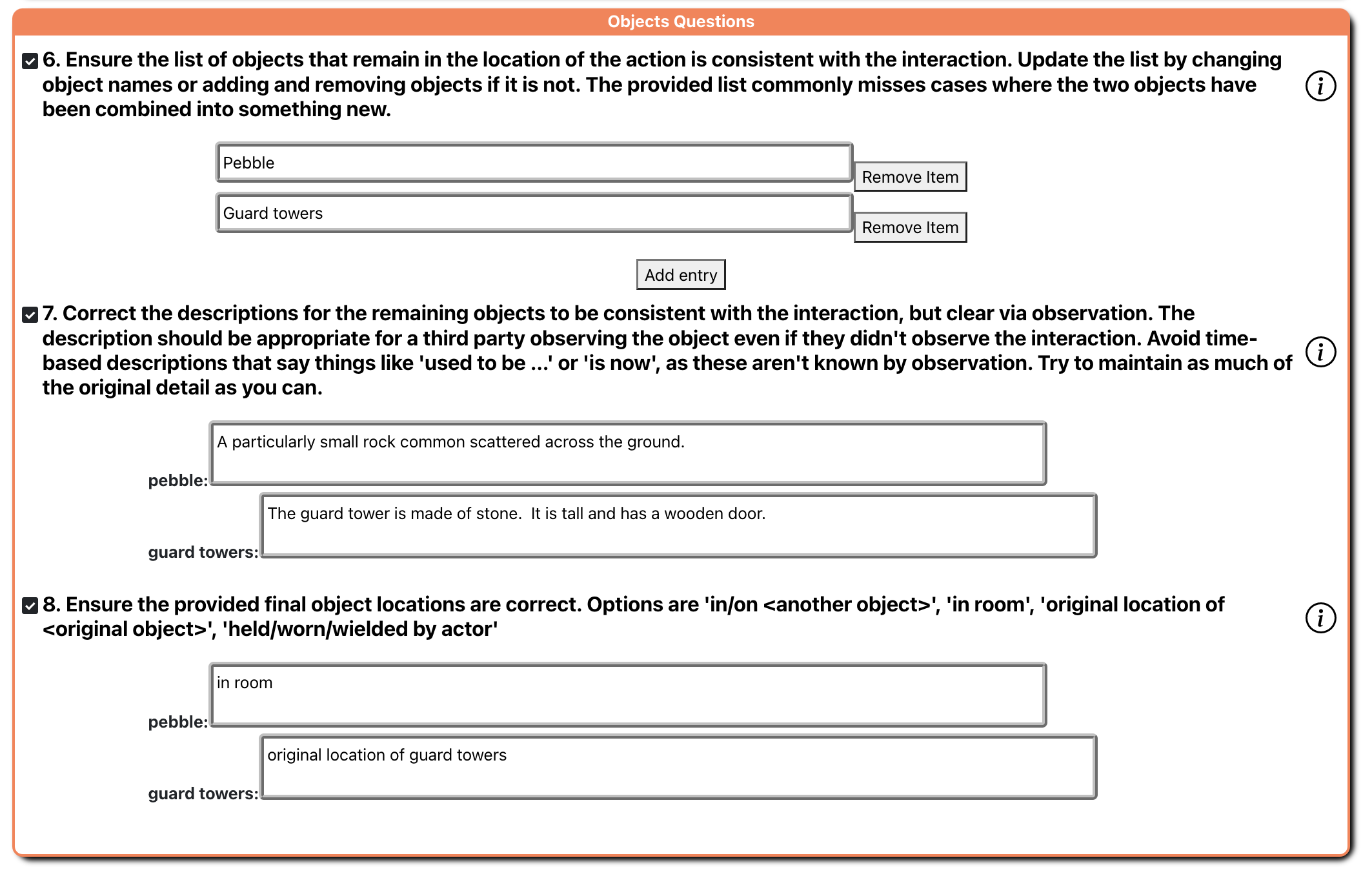}
    \caption{Interface to validate and correct the final objects,  their descriptions, and their locations.}
    \label{fig:grounding_objects}
\end{figure*}

They were then asked to confirm or alter the list of objects that would be present after completion of the interaction, and provide descriptions and locations for each of them. This is shown in Figure \ref{fig:grounding_objects}.

\begin{figure*}[ht]
    \centering
    \includegraphics[width=\textwidth]{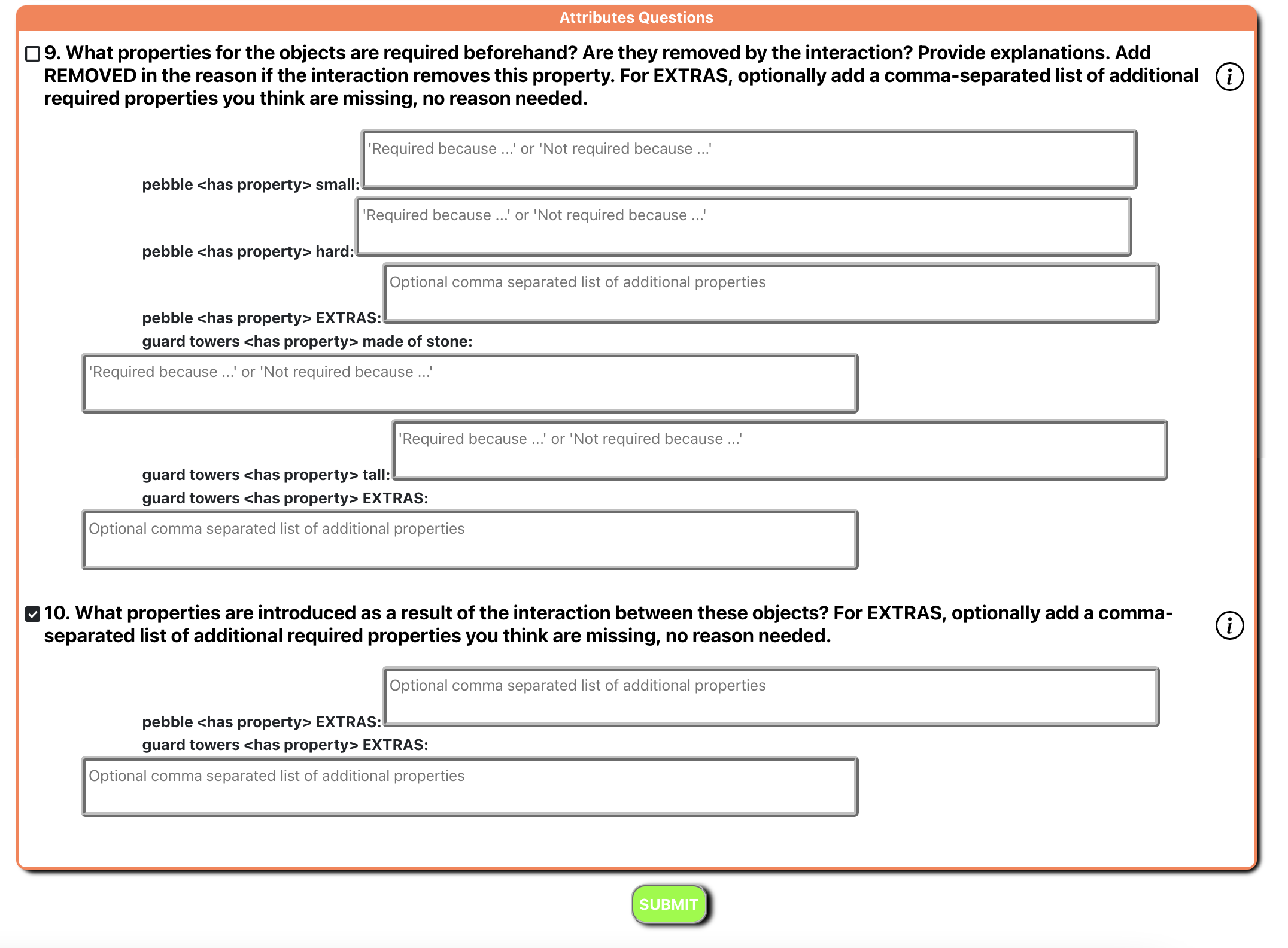}
    \caption{Interface to validate and correct the objects' required attributes before and after the interaction.}
    \label{fig:grounding_attributes}
\end{figure*}

Finally, they were asked to provide a list of attributes for each object that would be required for the interaction to occur, and a list of attributes that would be present after the interaction were to occur. This is shown in Figure \ref{fig:grounding_attributes}

After all of the tasks were collected and completed, we ran scripts to capture and reformat common data entry mistakes, then converted all of the events into a LIGHT-compatible format.

All of our collection interfaces were built and run using the Open-Source Mephisto framework \cite{mephisto}.

\section{Human Model Evaluation Methodology}
\label{sec:human-eval-method}
This section discusses the methodology for creating and running our human evaluation, the experimental details can be found in Section \ref{sec:human-eval}.

Our core task for running the human evaluations can be seen in Figure \ref{fig:human_eval}, and it derives from the "Model Chat" evaluation task present in ParlAI. The core difference here is we make a modification to how the context is passed along, as we are using a modified one-turn flow for interactions, while loading the context from pre-generated playthroughs rather than existing ParlAI tasks. 

Workers were selected from an allowlist of high-quality workers that had completed other tasks for this project.

\section{Additional \textsc{UseEvent} Examples}
\label{sec:more-use-events}

We also provide additional samples from the \textsc{UseEvent} dataset. Table \ref{tab:use_event2} is an interaction that creates a new object while modifying the previously existing ones. Table \ref{tab:use_event3} is an event that displays attribute changes. Table \ref{tab:use_event4} displays a simpler interaction that still makes significant graph updates, and provides an example where there's no alternate narration because the primary one was marked as invalid. Table \ref{tab:use_event5} displays a more intense interaction with an effective required attribute on the primary object. Table \ref{tab:use_event6} displays an interaction that is ``unsuccessful'', but still results in a graph update. Table \ref{tab:use_event_noop} provides an example that was collected using the ``boring'' variation of the task, and Table \ref{tab:use_event_failure_1} and Table \ref{tab:use_event_failure_2} provide examples for the ``Failed'' version of the task.

\begin{table*}[ht]
   \centering
   \begin{tabular}{|l |r | r |p{75mm}|}
   \hline
   \textbf{Action} & \textbf{Phrase} & \multicolumn{2}{p{95mm}|}{hook the rope around the chandelier}\\
   \cline{2-4}
   & \textbf{Narration} & \multicolumn{2}{p{95mm}|}{You wind up your arm holding the thick rope, pulling back with an incredible amount of momentum before you aim the circular hole in the rope towards the chandelier. You are amazed that you ensnare one of the edges of the chandelier. One of the candles falls off of the chandelier and crashes to the ground as the sound of chimes echo as the sparkly appendages of the chandelier crash against each other chaotically.}\\
   \cline{2-4}
   & \textbf{Alternate} & \multicolumn{2}{p{95mm}|}{You wind up and throw the thick rope's circular end towards the chandelier, ensnaring one of the chandelier's arms.  A candle falls off the chandelier, hitting the ground while sparkling arms of the chandelier clang together and ring out with chaotic noise.}\\
   \cline{2-4}
   & \textbf{External} & \multicolumn{2}{p{95mm}|}{\{actor\} winds up and throws the thick rope's circular end towards the chandelier. One of the chandelier's arms is ensnared.  A candle falls off the chandelier and noise rings out as the chandelier's dangling appendages clang into one another.}\\
   \hline
   \hline
   \textbf{Initial} & \textbf{Primary} & name & rope \\
   \cline{3-4}
   \textbf{Objects} & & description & Three thick strands made from lime bast are woven together to create a strong rope that can withstand almost any tension; a circle is tied at the end. \\
   \cline{3-4}
   & & Attributes & looped, strong \\
   \cline{2-4}
   & \textbf{Secondary} & name & chandelier \\
   \cline{3-4}
   & & description & The chandelier is magnificant, it is large and is made of iron, it had candles on every arm with some sort of sparkle that hangs down off of it. It is all you could look at. \\
   \hline
   \hline
   \textbf{Final} & \textbf{Name} & \multicolumn{2}{p{95mm}|}{rope} \\
   \cline{3-4}
   \textbf{Objects} & Description & \multicolumn{2}{p{95mm}|}{A rope made from lime bast that is strong and has a looped circle at the end. It is hooked around the chandelier.} \\
   \cline{3-4}
   & Location & \multicolumn{2}{p{95mm}|}{On chandelier} \\
   \cline{3-4}
   & Attribute Changes & \multicolumn{2}{p{95mm}|}{+Entangled} \\
   \cline{2-4}
   & \textbf{Name} & \multicolumn{2}{p{95mm}|}{chandelier} \\
   \cline{3-4}
   & Description & \multicolumn{2}{p{95mm}|}{A large, iron chandelier with candles and sparkly appendages. It is swinging and attached to a rope.} \\
   \cline{3-4}
   & Location & \multicolumn{2}{p{95mm}|}{Original location of chandelier} \\
   \cline{3-4}
   & Attribute Changes & \multicolumn{2}{p{95mm}|}{+Incomplete, +Unbalanced} \\
   \cline{2-4}
   & \textbf{Name} & \multicolumn{2}{p{95mm}|}{candle} \\
   \cline{3-4}
   & Description & \multicolumn{2}{p{95mm}|}{A candle from a chandelier} \\
   \cline{3-4}
   & Location & \multicolumn{2}{p{95mm}|}{In room} \\
   \hline

   \end{tabular}
   \caption{Example from the \textsc{UseEvents} dataset}
   \label{tab:use_event2}
\end{table*}

\begin{table*}[ht]
   \centering
   \begin{tabular}{|l |r | r |p{75mm}|}
   \hline
   \textbf{Action} & \textbf{Phrase} & \multicolumn{2}{p{95mm}|}{catch tropical bird with casting net}\\
   \cline{2-4}
   & \textbf{Narration} & \multicolumn{2}{p{95mm}|}{You cast the net in the direction of the bright colored bird.  The bird attempts to flee but flies right into the net. The steel balls hold the net to the ground trapping the tropical bird underneath.   The bird lays panting with wings spread and does not fight the capture.}\\
   \cline{2-4}
   & \textbf{Alternate} & \multicolumn{2}{p{95mm}|}{You cast the net at the tropical bird. The bird attempts to flee but flies right into the net, becoming ensnared. The bird lies panting with its wings spread and does not attempt to resist the capture.}\\
   \cline{2-4}
   & \textbf{External} & \multicolumn{2}{p{95mm}|}{\{actor\} casts a net in the direction of a brightly colored bird. The bird flies into the net and is trapped.}\\
   \hline
   \hline
   \textbf{Initial} & \textbf{Primary} & name & casting net \\
   \cline{3-4}
   \textbf{Objects} & & description & A woven matrix of ropes with small balls of weighted steel on the ends.  Perfect for casting into the water for fishing or tangling up into a birds nest for trapping. \\
   \cline{2-4}
   & \textbf{Secondary} & name & tropical bird \\
   \cline{3-4}
   & & description & The tropical bird is brightly colored with a simple pattern. \\
   \cline{3-4}
   & & Attributes & calm \\
   \hline
   \hline
   \textbf{Final} & \textbf{Name} & \multicolumn{2}{p{95mm}|}{Trapped bird} \\
   \cline{3-4}
   \textbf{Objects} & Description & \multicolumn{2}{p{95mm}|}{The tropical bird is brightly colored with a simple pattern. It is caught in a casting net.} \\
   \cline{3-4}
   & Location & \multicolumn{2}{p{95mm}|}{in casting net} \\
   \cline{3-4}
   & Attribute Changes & \multicolumn{2}{p{95mm}|}{-calm, +trapped} \\
   \cline{2-4}
   & \textbf{Name} & \multicolumn{2}{p{95mm}|}{casting net} \\
   \cline{3-4}
   & Description & \multicolumn{2}{p{95mm}|}{A woven matrix of ropes with small balls of weighted steel on the ends. A tropical bird is ensnared in it.} \\
   \cline{3-4}
   & Location & \multicolumn{2}{p{95mm}|}{In room} \\
   \cline{3-4}
   & Attribute Changes & \multicolumn{2}{p{95mm}|}{+tangled} \\
   \hline

   \end{tabular}
   \caption{Example from the \textsc{UseEvents} dataset}
   \label{tab:use_event3}
\end{table*}

\begin{table*}[ht]
   \centering
   \begin{tabular}{|l |r | r |p{75mm}|}
   \hline
   \textbf{Action} & \textbf{Phrase} & \multicolumn{2}{p{95mm}|}{Add mead to the pitcher}\\
   \cline{2-4}
   & \textbf{Narration} & \multicolumn{2}{p{95mm}|}{You fill the pitcher with mead, making sure to take in the sweet aroma as it flows into the pitcher.}\\
   \cline{2-4}
   & \textbf{Alternate} & \multicolumn{2}{p{95mm}|}{N/A}\\
   \cline{2-4}
   & \textbf{External} & \multicolumn{2}{p{95mm}|}{\{actor\} fills their pitcher with mead.}\\
   \hline
   \hline
   \textbf{Initial} & \textbf{Primary} & name & pitcher \\
   \cline{3-4}
   \textbf{Objects} & & description & The pitcher is heavier than it looks, and appears to still be in one piece. \\   \cline{3-4}
   & & Attributes & empty \\
   \cline{2-4}
   & \textbf{Secondary} & name & Mead \\
   \cline{3-4}
   & & description & An alcoholic brew made from fermented honey and spices.  It's a sweet mixture that lightens the spirit.  The mead is usually kept in a large keg barrel at the local tavern \\
   \hline
   \hline
   \textbf{Final} & \textbf{Name} & \multicolumn{2}{p{95mm}|}{Mead} \\
   \cline{3-4}
   \textbf{Objects} & Description & \multicolumn{2}{p{95mm}|}{A sweet alcoholic brew made from fermented honey and spices.} \\
   \cline{3-4}
   & Location & \multicolumn{2}{p{95mm}|}{in pitcher} \\
   \cline{2-4}
   & \textbf{Name} & \multicolumn{2}{p{95mm}|}{pitcher} \\
   \cline{3-4}
   & Description & \multicolumn{2}{p{95mm}|}{A container for holding liquids. You can hear a something sloshing around in it as you pick it up.} \\
   \cline{3-4}
   & Location & \multicolumn{2}{p{95mm}|}{Held by \{actor\}} \\
   \cline{3-4}
   & Attribute Changes & \multicolumn{2}{p{95mm}|}{+full} \\
   \hline

   \end{tabular}
   \caption{Example from the \textsc{UseEvents} dataset. Not all entries have an alternate description as some were filtered for quality reasons.}
   \label{tab:use_event4}
\end{table*}

\begin{table*}[ht]
   \centering
   \begin{tabular}{|l |r | r |p{75mm}|}
   \hline
   \textbf{Action} & \textbf{Phrase} & \multicolumn{2}{p{95mm}|}{Burn the tents with the lit torch}\\
   \cline{2-4}
   & \textbf{Narration} & \multicolumn{2}{p{95mm}|}{You stand close to the tents and lean in with the lit torch.  The tents catch fire.  They start to burn and you just watch.}\\
   \cline{2-4}
   & \textbf{Alternate} & \multicolumn{2}{p{95mm}|}{You use the torch to set the tents on fire by standing close to them, leaning in with the lit torch, and allowing them to begin to burn.}\\
   \cline{2-4}
   & \textbf{External} & \multicolumn{2}{p{95mm}|}{\{actor\} stands close to some tents with a lit torch in hand. The tents catch fire and {actor} just seems to watch.}\\
   \hline
   \hline
   \textbf{Initial} & \textbf{Primary} & name & lit torch \\
   \cline{3-4}
   \textbf{Objects} & & description & The lit torch is in your hand.  It is burning furiously. \\  
   \cline{3-4}
   & & Attributes & burning \\ 
   \cline{2-4}
   & \textbf{Secondary} & name & empty tent \\
   \cline{3-4}
   & & description & The tent looks small, as though only one or two could fit comfortably inside it.  It is old and weathered, but seems sturdy despite its age. \\
   \cline{3-4}
   & & Attributes & flammable \\
   \hline
   \hline
   \textbf{Final} & \textbf{Name} & \multicolumn{2}{p{95mm}|}{lit torch} \\
   \cline{3-4}
   \textbf{Objects} & Description & \multicolumn{2}{p{95mm}|}{The lit torch is burning furiously.} \\
   \cline{3-4}
   & Location & \multicolumn{2}{p{95mm}|}{Held by \{actor\}} \\
   \cline{2-4}
   & \textbf{Name} & \multicolumn{2}{p{95mm}|}{flaming tent} \\
   \cline{3-4}
   & Description & \multicolumn{2}{p{95mm}|}{The tent is partially on fire. The parts that aren't appear old and weathered.} \\
   \cline{3-4}
   & Location & \multicolumn{2}{p{95mm}|}{original location of empty tent} \\
   \cline{3-4}
   & Attribute Changes & \multicolumn{2}{p{95mm}|}{+ablaze} \\
   \hline

   \end{tabular}
   \caption{Example from the \textsc{UseEvents} dataset.}
   \label{tab:use_event5}
\end{table*}

\begin{table*}[ht]
   \centering
   \begin{tabular}{|l |r | r |p{75mm}|}
   \hline
   \textbf{Action} & \textbf{Phrase} & \multicolumn{2}{p{95mm}|}{wrap the bunny in the rabbit fur coverlet}\\
   \cline{2-4}
   & \textbf{Narration} & \multicolumn{2}{p{95mm}|}{You attempt to wrap the bunny in the rabbit fur coverlet, but apparently, the bunny thinks this is in bad taste and hops away.}\\
   \cline{2-4}
   & \textbf{Alternate} & \multicolumn{2}{p{95mm}|}{You attempt to wrap the bunny in the rabbit fur coverlet, but the bunny hops away.}\\
   \cline{2-4}
   & \textbf{External} & \multicolumn{2}{p{95mm}|}{\{actor\} tries to wrap a bunny in a rabbit fur coverlet, but the bunny escapes.}\\
   \hline
   \hline
   \textbf{Initial} & \textbf{Primary} & name & rabbit fur coverlet \\
   \cline{3-4}
   \textbf{Objects} & & description & The rabbit fur coverlet looks very warm and soft. \\
   \cline{2-4}
   & \textbf{Secondary} & name & bunny \\
   \cline{3-4}
   & & description & A precious little bunny rabbit with long ears. \\
   \hline
   \hline
   \textbf{Final} & \textbf{Name} & \multicolumn{2}{p{95mm}|}{rabbit fur coverlet} \\
   \cline{3-4}
   \textbf{Objects} & Description & \multicolumn{2}{p{95mm}|}{The rabbit fur coverlet looks very warm and soft. It lies on the ground, abandoned.} \\
   \cline{3-4}
   & Location & \multicolumn{2}{p{95mm}|}{In room} \\
   \cline{3-4}
   & Attribute Changes & \multicolumn{2}{p{95mm}|}{+Wearable} \\
   \hline

   \end{tabular}
   \caption{Example from the \textsc{UseEvents} dataset. Not all events play the suggested phrase out successfully}
   \label{tab:use_event6}
\end{table*}

\begin{table*}[ht]
   \centering
   \begin{tabular}{|l |r | r |p{75mm}|}
   \hline
   \textbf{Action} & \textbf{Phrase} & \multicolumn{2}{p{95mm}|}{Melt the bells in the pot of stew}\\
   \cline{2-4}
   & \textbf{Narration} & \multicolumn{2}{p{95mm}|}{You dip one of the bells into the pot of stew as a test. The stew isn't hot enough to melt it, which is probably a good thing for anyone who might want to eat it.}\\
   \cline{2-4}
   & \textbf{Alternate} & \multicolumn{2}{p{95mm}|}{N/A}\\
   \cline{2-4}
   & \textbf{External} & \multicolumn{2}{p{95mm}|}{\{actor\} dips a bell into a pot of stew, but it doesn't seem to melt.}\\
   \hline
   \hline
   \textbf{Initial} & \textbf{Primary} & name & bells \\
   \cline{3-4}
   \textbf{Objects} & & description & The bells are loud and shiny silver. \\  
   \cline{2-4}
   & \textbf{Secondary} & name & pot of stew \\
   \cline{3-4}
   & & description & A pot of stew filled with chunks of lamb and garden fresh vegetables. It's a little salty. \\
   \hline

   \end{tabular}
   \caption{`Boring' example from the \textsc{UseEvents} dataset. Entries like these have no final state, as there is no change.}
   \label{tab:use_event_noop}
\end{table*}

\begin{table*}[ht]
   \centering
   \begin{tabular}{|l |r | r |p{75mm}|}
   \hline
   \textbf{Action} & \textbf{Phrase} & \multicolumn{2}{p{95mm}|}{carve the big ornate door with the butter}\\
   \cline{2-4}
   & \textbf{Narration} & \multicolumn{2}{p{95mm}|}{You cannot carve the big ornate door by using the butter, the butter is so soft and can't leave the smallest dent on the door, you decide not to waste the butter trying further.}\\
   \cline{2-4}
   & \textbf{Alternate} & \multicolumn{2}{p{95mm}|}{N/A}\\
   \cline{2-4}
   & \textbf{External} & \multicolumn{2}{p{95mm}|}{\{actor\} tries to carve the big ornate door with the butter, but fails as the butter is too soft to make any mark.}\\
   \hline
   \hline
   \textbf{Initial} & \textbf{Primary} & name & butter \\
   \cline{3-4}
   \textbf{Objects} & & description & the butter is soft and delicious \\  
   \cline{2-4}
   & \textbf{Secondary} & name & big ornate doors \\
   \cline{3-4}
   & & description & The door is made of a beautiful, bright wood, and it is covered in carvings of trees and animals. \\
   \hline

   \end{tabular}
   \caption{`Failed' example from the \textsc{UseEvents} dataset. Entries like these have no final state, as there's no change.}
   \label{tab:use_event_failure_1}
\end{table*}

\begin{table*}[ht]
   \centering
   \begin{tabular}{|l |r | r |p{75mm}|}
   \hline
   \textbf{Action} & \textbf{Phrase} & \multicolumn{2}{p{95mm}|}{Pull down the 80-foot tall bronze statue of an 8 legged goddess with the twine}\\
   \cline{2-4}
   & \textbf{Narration} & \multicolumn{2}{p{95mm}|}{You cannot pull down the 80-foot tall bronze statue of an 8 legged goddess with just twine. It is not strong or long enough. It won't do anything if you try.}\\
   \cline{2-4}
   & \textbf{Alternate} & \multicolumn{2}{p{95mm}|}{N/A}\\
   \cline{2-4}
   & \textbf{External} & \multicolumn{2}{p{95mm}|}{\{actor\} tried to pull down an 80-foot tall bronze statue of an 8 legged goddess with a piece of twine, but it wasn't strong or long enough and didn't do anything.}\\
   \hline
   \hline
   \textbf{Initial} & \textbf{Primary} & name & twine \\
   \cline{3-4}
   \textbf{Objects} & & description & The twine is a foot in length and fairly thin \\  
   \cline{2-4}
   & \textbf{Secondary} & name & 80-foot tall bronze statue of an 8 legged goddess \\
   \cline{3-4}
   & & description & The deity seems to stare into the distance, the mere mortals bellow her not worthy of her gaze. \\
   \hline

   \end{tabular}
   \caption{`Failed' example from the \textsc{UseEvents} dataset. Entries like these have no final state, as there's no change.}
   \label{tab:use_event_failure_2}
\end{table*}

\end{document}